# NER-Luxury: Named entity recognition for the fashion and luxury domain

*Akim Mousterou (A-M Research)*
moakim@protonmail.com

***Abstract***
In this study, we address multiple challenges of developing a named-entity recognition model in English for the fashion and luxury industry, namely the entity disambiguation, French technical jargon in multiple sub-sectors, scarcity of the ESG methodology, and a disparate company structures of the sector with small and medium-sized luxury houses to large conglomerate leveraging economy of scale.

In this work, we introduce a taxonomy of 36+ entity types with a luxury-oriented annotation scheme, and create a dataset of more than 40K sentences respecting a clear hierarchical classification. We also present five supervised fine-tuned models NER-Luxury for fashion, beauty, watches, jewelry, fragrances, cosmetics, and overall luxury, focusing equally on the aesthetic side and the quantitative side.

In an additional experiment, we compare in a quantitative empirical assessment of the NER performance of our models against the state-of-the-art open-source large language models that show promising results and highlights the benefits of incorporating a bespoke NER model in existing machine learning pipelines.

## 1. Introduction

From artistry to political economy, philosophers of Ancient Greece already discussed the meanings and ramifications of the idea of luxury (Berry, 1994). Over the last several decades, the luxury industry has morphed into a global market, one of the most valuable sectors in France, and an important sector in Europe. Nevertheless, based on aesthetic values of artistic directors, this sector has been difficult to map network effects, to quantify relevant signals, and understand optimal strategies.

For many years, economists, theorists and scholars have been passionate about the pricing of luxury goods based on scarcity (Smith, 1776), on the mechanism of value according to wealthy buyers (Ricardo, 1817) (Marshall, 1890), on the social aspect of consuming luxury goods (Veblen, 1899), and on the psychological effects such as the scarcity principle, formalized in the "*Commodity theory*" (Brock, 1968).

The economic theory of "*Design Innovation and Fashion cycles*" (Pesendorfer, 1995) and the response "*Fashion Cycles in Economics*" (Coelho et al., 2004) brings those observations to the economic field by quantifying the complex buyer interactions and the importance of branding, over the quality of raw materials, or craftsmanship. Similarly, in the socioeconomic sphere, Jean Baudrillard explained that in post-industrial societies "*Sign value*" (Baudrillard, 1968) has surpassed the other economic values based on production cost, and pure market value.

To understand the value of luxury goods from a consumer perspective in 2024, "*the Distinction*" (Bourdieu, 1979), the sociology research on the cartography of social structure to understand logic of taste are no longer relevant due to the complexity of modern consumer paths, with the power of network effects with social media platforms (Rohlfs, 1974), the digital identity at the age of *hyperreality* (Baurdillard, 1981), and the luxury goods, as an asset class for investment strategy.

From an M&A transactions perspective, with many intangible assets, the difficult valuation of luxury houses could be linked to the desire-satisfaction theory of value theorized in *Ethics* (Spinoza, 1677), where the value of an acquisition is the desire of the buyer, whatever leads to the satisfaction of the buyer. Luxury houses are priced rather than valued and could be classified as "*Trophy assets*" (Damodaran, 2023).

Executives, strategists, analysts, researchers, and even individual investors can be overwhelmed by the abundance of positive and negative signals from the aesthetic side and the quantitative side, primarily coming from news source, specialized websites, or regulated financial reports.

In order to understand the underlining mechanisms, and push the boundaries of quantitative research, we will create a Named Entity Recognition (NER) model to extract relevant information from unstructured texts. A crucial task of natural language processing, NER aims to detect segments of text that correspond to named entities and categorize them into predefined groups like person (PER), location (LOC), or organization (ORG).

## 2. Related work: adaptation and application

### *2.1 Entity disambiguation*

In the luxury domain, NER can help disambiguate entities with the same name according to the context. For example, *Louis Vuitton* can refer to the visionary French founder (1821 – 1892), the luxury house, the multiples companies, the legal entity, the French luxury listed-group, the various collections, the art museum, the sailing competition, the city guides, or the iconic leather goods.

NER can be used to extract for information extraction and retrieval, for example, which luxury houses were cited by the Chairman and CEO of *Kering*, François-Henri Pinault in the latest shareholder letter's. In the age of social media, NER can be used to identify and monitor the impact of a celebrity, a blogger, or an event during a marketing campaign social media platforms such as *Instagram, Youtube, LinkedIn, WeChat, LINE*, etc.

### **2.2 Related work**

To the best of our knowledge, no NER model specialized for the luxury domain has been reported in literature. Researchers at the e-Commerce giant, *eBay* and online luxury fashion platform, *Farfetch*, have been focus on combining NER pipelines with Knowledge Graph to increase the accuracy of their product textual descriptions or extractions on a restricted corpus dedicated to their respective solutions (Putthividhya and Hu, 2011) (Zhou et al., 2021) (Barroca et al., 2022).

Moreover, the fashion and luxury domain presents specific challenges in defining clear lines between fast-fashion and luxury, that are opposite, if not orthogonal according to their business models, pricing strategy, supply-chain management, and sustainability practices.

NER models adapted to the medical and legal domain have been leading the NER methodology in Natural Language Processing. Recently, we have witnessed the creation of domain-specific NER models such as astronomy (Evans et al., 2024), historical linguistics (Avram et al., 2024), global event (Efeoglu et al., 2024), and fantasy gaming (Sivaganeshan et al., 2024).

Therefore, in this present study, we contribute of this paper as follows:

1. Create an empirical scheme of the luxury industry based on a financial index
2. Build an large annotation dataset covering our research-domain
3. Fine-tuned various NER models to our bespoke dataset
4. Evaluated the NER models to Llama 3.1 models

## *3. Methodology*

**3.1 Hierarchical scheme**

We will be grounding our methodology based on a financial index to incorporate the majority of prestigious houses from luxury groups, and then create a hierarchical classification to separate luxury houses, from brands, from fast-fashion brands with multiple sustainability indexes.

We began by sourcing financial framework, the *S&P Global Luxury Index* has out of the scope constituents from the provision of luxury services, such as *Marriott Intl A, Royal Caribbean Group*, and *Tesla Inc*. While, the *MSCI Europe Textiles, Apparel & Luxury Goods Index* focus exclusively in the European market. Therefore, we will leverage an updated version of the **Savigny Luxury Index** ("**SLI**") created in the early 2000 by the former managing director M&A boutique, *Savigny Partners* and actual Co-Head of Merchant Banking at *Stanhope Capital*, Pierre Mallevays.

The *SLI index* comprises of a basket of carefully handpicked luxury goods manufacturers, and retailers. As of 2023, the top 5 companies in the SLI account for close to 80 percent of the total sales generated by the 17 companies tracked by the SLI, and they own 142 out of the 190 brands covered by the index (Mallevays, 2023). We can complete the *SLI index* by adding important non-listed groups, *Chanel limited, Rolex SA, Comme des Garçons, Giorgio Armani SpA, Patek Philippe SA, OTB SpA*, and recently listed groups such as *Lanvin Group, Puig S.L.,* and *Ermenegildo Zegna Group*.

**3.2 Sustainability**

In the fashion industry, *MSI* and the *Higg Index* are the most used available datasets for developing robust global greenhouse gas (GHG) emissions by the fashion industry. Various ESG experts and non-profit organizations around the world have criticized the sustainability rating systems (Sadowksi, et al., 2021).

In the Anthropocene era, the transparency in fashion can be an important factor in the brand perception for consumers but doesn't reveal robust sustainable practices. The most polluting fast-fashion brands are on the top of *Fashion Transparency index* (Jestratijevic et al., 2021). Recent experimental research leveraging online data to separate fast-fashion brands from premium brands by calculating the percentage of synthetic fibers against natural fibers in a garment scaled by the price. In addition, this creative methodology, various premium brands or innovative fashion designers using synthetic fibers for technical necessities can be discarded by the number of SKU in a collection or the retail space productivity.

**3.3 Dataset building**

We built a custom dataset by extracting each major entity of luxury groups, and their extended universe from Wikipedia (CC BY-SA). Despite the powerful historical approach of Wikipedia, we observed many mistakes on the types of corporate legal entities and governing bodies. Therefore, Executive Committee and Board of Directors of fashion and luxury groups were added according to regulated information that can be found publicly in annual financial reports such as Universal Registration Document enforced by the *Autorité des marchés financiers* (*A.M.F*) in France or the Proxy Statements (Schedule 14A information) enforced by the *U.S. Securities and Exchange Commission* (*S.E.C*) in the United States of America. In each paragraph with textual entailment, we manually replaced pronoun references to its referent entities according to the context to reduce ambiguous sentences, and augment the number of entities.

## *3.4 Labels of entities*

In total, more than 41,046 sentences are split into a training and a test set, containing 36,941 sentences for training, and 4105 sentences for the test. We leveraged the B-I-O annotation scheme (Ramshaw and Marcus, 1999), where each word in a sentence is associated with one tag. The "B" tag indicates that the associated word marks the Beginning of a span; "I" indicates a word Inside a span, and "O" marks words Outside a span. Entities were manually annotated internally at AM Research using open-source text annotation tool, ***Doccano***.

The annotation started in December 2022. The number of labels evolved rapidly from July 2023 to September 2023. As of July 2024, we believe it is a strong basis for the taxonomy to build on, while also being aware of the quick evolution of the fashion and luxury industry with constant nominations, events, or, M&A activities. We proposed the following taxonomy for named-entity recognition to identify major entities, map intermediate and minor bodies, while estimate their relevancy.

*Labels of NER Luxury*

| Label | Description and example |
|---|---|
| **O** | *Outside* (of a text segment) |
| **Date** | *Temporal expressions* (1854, Q2 2023, Nineties, September 21) |
| **Location** | *Physical location and area* (Paris, Japan, Europe, Champs-Elysées) |
| **Event** | *Critical events* (WW II, Olympics, IPO, Covid pandemic, Paris Fashion Week) |
| **MonetaryValue** | *Currency, price, sales, revenue* ($2.65 billion, 4.6 million euros, CHF 400,000, etc.) |
| **House** | *Fashion and luxury houses* (Louis Vuitton, Cartier, Gucci, Chanel) |
| **Brand** | *Sportswear, beauty and labels* (Nike, Lululemon, Clinique) |
| **FastFashion** | *Mass-market retailers* (Zara, H&M, Uniqlo, Shein) |
| **PrivateCompany** | *Unlisted companies* (Chanel SA, Stella McCartney Ltd, Valentino S.p.A) |
| **ListedGroup** | *Listed groups* (LVMH, Hermès International SCA, Kering) |
| **HoldingTrust** | *Holding and family office* (Agache, H51, Mousse Partners, Artèmis) |
| **InvestmentFirm** | *Investment banks, PE funds, M&A firms* (KKR, L Catterton, Mayhoola, Bernstein) |
| **MediaPublisher** | *Media outlets* (Bloomberg, Vogue, Business of Fashion, NYT) |
| **Hospitality** | *Luxury hospitality* (Ritz Paris, Belmond hotel Cipriani, Venetian Macao) |
| **MuseumGallery** | *Exhibition spaces* (Louvre, MET, Victoria & Albert, Pinault Collection) |
| **Retailer** | *POS, department stores, and select shops* (Bergdorf, Le Bon Marché, Takashimaya) |
| **Education** | *Business and fashion schools* (Polytechnic, Harvard, LSE, ESCP, Central Saint Martins, IFM) |
| **Organization** | *Legal, scientific, and cultural entities* (CFDA, European Union, UNESCO, SEC) |
| **ArtisticDirector** | *Lead creative of houses* (Karl Lagerfeld, Daniel Lee, Sarah Burton, Alessandro Michele) |
| **Executive** | *C-level, board members* (Jérôme Lambert, Sue Nabi, Pietro Beccari) |
| **Founder** | *Founder, creative, and owner* (Ralph Lauren, Rei Kawakubo, Michael Kors) |
| **Chairperson** | *Chairman/Chairwoman* (e.g. Bernard Arnault, Patrizio Bertelli, François-Henri Pinault) |
| **AnalystBanker** | *Equity analysts, M&A bankers* (Luca Solca, Pierre Mallevays, Louise Singlehurst) |
| **KOL** Key Opinion Leader | *Artists, celebrities, historical figures* (Audrey Hepburn, BTS, Kanye West, Emma Watson) |
| **AthleteTeam** | *Professional athletes, and teams* (David Beckham, Serena Williams, Luna Rossa) |
| **Model** | *Fashion models* (Iman, Kate Moss, Adriana Lima, Naomi Campbell, Mariacarla Boscono) |
| **CreativeInsider** | *Photographers, make-up artists, watchmakers* (Nick Knight, Dominique Ropion, Gérald Genta) |
| **EditorJournalist** | *Editor-in-chief, fashion editors, journalists* (Suzy Menkes, Anna Wintour, Carine Roitfeld) |
| **GarmCollection** | *Iconic garment and collections* (Haute Couture, Bar suit, No.13 of McQueen, Jungle Dress) |
| **Cosmetic** | *Cosmetic products* (Tilbury Glow palette, Crème de La Mer, YSL Nu, Viva Glam) |
| **Fragrance** | *Perfumes, and EdT* (Chanel No.5, Dior Sauvage, Terre d'Hermès, Tom Ford Black Orchid) |
| **BagTrvlGoods** | *Bags, and leather goods* (Hermès Birkin bag, Louis Vuitton Speedy bag, Chanel 2.55) |
| **Jewelry** | *Fine jewelry, and gems* (Alhambra of Van Cleef & Arpels, Juste un Clou Cartier, Winston Blue) |

| Timepiece | *Fine watches* (Nautilus Patek Philippe, Reverso Jaeger-Lecoultre, Rolex Oyster) |
|---|---|
| Footwear | *High heels to sneakers* (Rainbow of Ferragamo, Armadillo of McQueen, Air Force1) |
| WineSpirit | *Wine and spirit* (Château d'Yquem, Clos de Tart, Château Matras, Hennessy, Moet, Belvedere) |
| Sustainability | *Relevant ESG factors and entities* (Ethical Fashion Initiative, decoupling, biodiversity loss) |
| CulturalArtifact | *Songs, books, movies* (The Devil wears Prada, American Gigolo, The College Dropout) |

## 5. Observations and biases

### 5.1 French influence

The lexicon in the fashion and luxury industry is strongly biased toward French technical words that are clearly out-of-vocabulary (OOV) in English. We witnessed the dominance of French linguistic features with accentuation for example *Hermès*, prêt-à-porter, *Château-d'Yquem, Comme des Garçons*, or Héliotrope from multiple sub-sectors, but also other languages such as *Azzedine Alaïa* in Latin script, FEИTY in Cyrillic script, *Dsquared²* in mathematical superscript expressions, or *A. Lange & Söhne* with the letter ö from the German alphabet.

### 5.2 Axiological neutrality

During classification, we made the deliberate choice of labeling luxury watchmakers such as *Cartier, Patek Philippe, A. Lange & Soehne*, etc. with the label "House" for their duality in creativity, and technicality. We also classified the perfume houses, such as *Diptyque, Guerlain, Lancôme*, etc. with the label "House" for their creative side. While, we attributed the label "Brand" to cosmetic brands such as *la Roche-Posay, Clinique*, and *La Mer* due to scientific formulation and strict regulatory requirements.

### 5.3 Label distributions

In the label distributions, the most common label is logically "O" (outside) with 511,047 entities. Then due to historical nature of Wikipedia, general labels such as "Date" with 21,847 entities and "Location" with 13,277 entities.

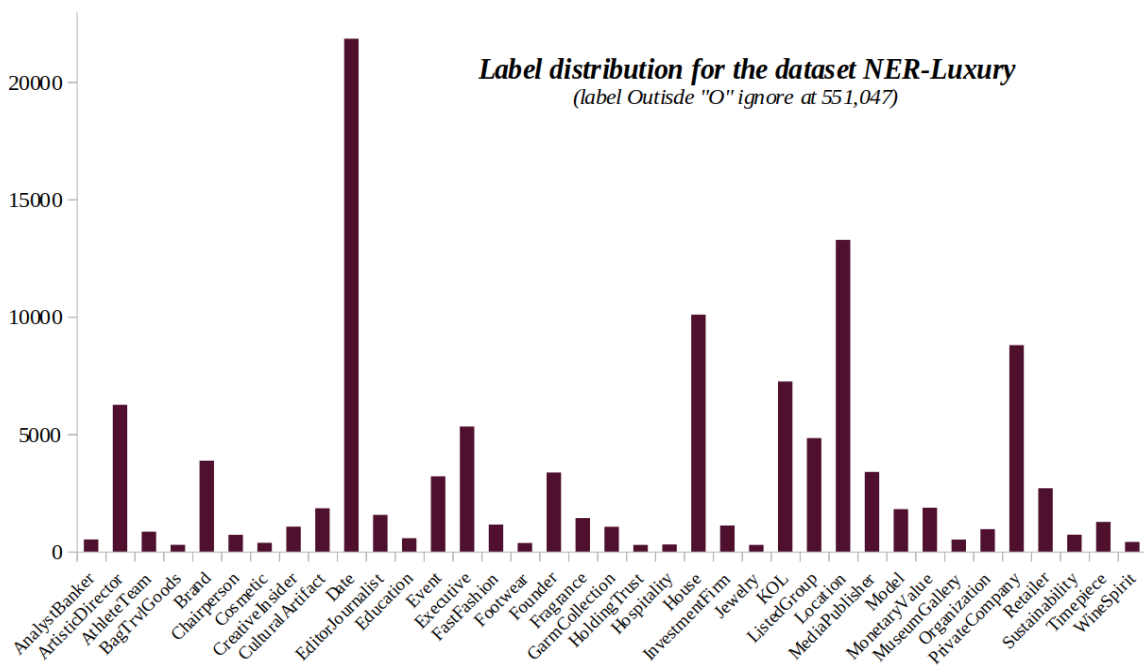

*Figure 1: Label distribution for the dataset Ner-Luxury - Akim Mousterou, 2024*

Our bespoke methodology for the luxury industry, we witnessed the heavily influence with the label "House" with more than 10,091 entities and "PrivateCompany" with more than 8,791 entities that have been consolidated or disappeared over the years. The dataset is naturally unbalanced in the portfolio of houses for listed groups, *LVMH Group* is composed of more than 75 houses (*Louis Vuitton, Dior, Tiffany, Fendi, etc.*), while *Burberry* is one major fashion house.

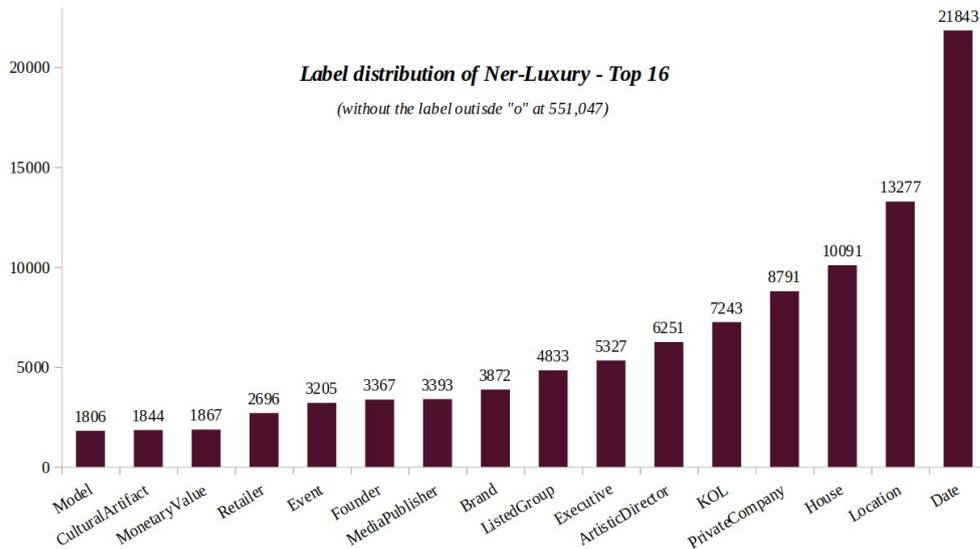

Figure 2: *Top 16 of labels in the Ner-Luxury dataset - Akim Mousterou, 2024*

The legal entities with label "HoldingTrust" such as *Financière Agache* of *LVMH SE*, *Artémis Group* of *Kering* or *Mousse Partners* of *Chanel*, etc., are quite rare wit 286 entities. Similarly, the presiding officer entities with the label "Chairperson" such as Bernard Arnault of *LVMH*, François-Henri Pinault of *Kering*, or Alain Weithmer of *Chanel*, etc., are rare but omnipresent both in the quantitative side and the aesthetic side due to their strategic position and responsibility for driving its long-term value creation. On the contrary, at 5,327 entities, the label "Executive" refers to c-level executives playing a crucial role in the functioning of the company with short-term incentives, benefits, and sometimes equity options.

*5.4 Temporal lineage*

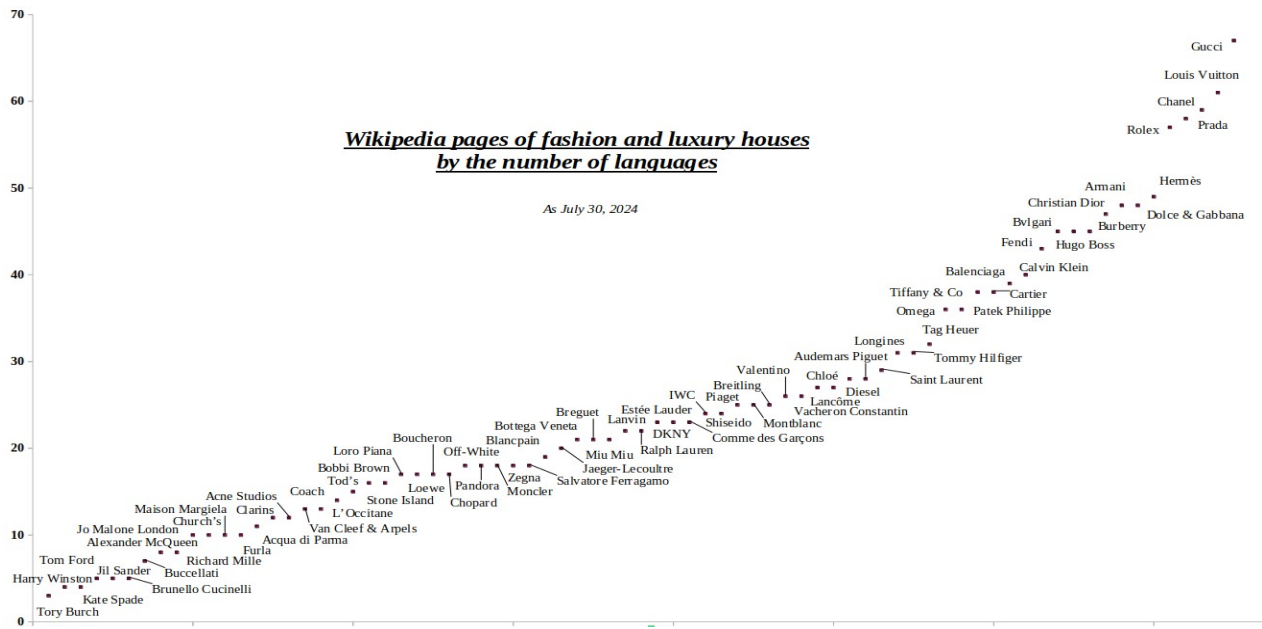

Figure 3: *Wikipedia pages fashion and luxury houses by number of languages - Akim Mousterou, 2024*

On figure 3, we can view the global awareness of luxury houses by plotting the number of languages on Wikipedia. We can also observe that the most performing houses in revenues have highest number of languages, due to the prestigious heritage built and maintained over time, and their global expansion.

The actual leadership position of the luxury conglomerate, *LVMH* has been achieved by m&a activities led by Bernard Arnault in the late 1990s and early 2000s. While over the years, *Hermès*' expansion was based on the uniqueness of the French luxury House. Controlled by the three main families, *Hermès* evolved from 3 descendants in the fourth generation to more than 100 members in the seventh generations, Europe's richest, with a combined fortune this year of about $151 billion.

Regardless of the corporate size, the development a luxury brand takes time and cannot be achieved overnight, therefore the stability of a family-owned group is the optimal form in the luxury industry. Therefore, the descendants of the "Chairperson" have automatically been granted the "Executive" label.

## 5.5 Artistic hegemony

The crucial role of founder with the label "Founder" can regroup various roles, the leading role for the aesthetic vision as "ArtisticDirector", in charge of the management and business development as "Executive", and the "Chairperson" role with total or partial ownership of the company.

On Wikipedia, we witnessed that visionary founders that started around the 1970s (e.g, *Ralph Lauren, Jean-Paul Gaultier*) have their legacy entirely based on them, the fashion and luxury houses are explained through the eyes of their founders. Due to the importance of the founders, their fashion houses doesn't even have a Wikipedia page (e.g. *Paco Rabanne, Dries Van Noten*). During the succession plan, the transition from the founder to a new artistic director brings instability on both the creative side and the business side for the legacy of the house or the company.

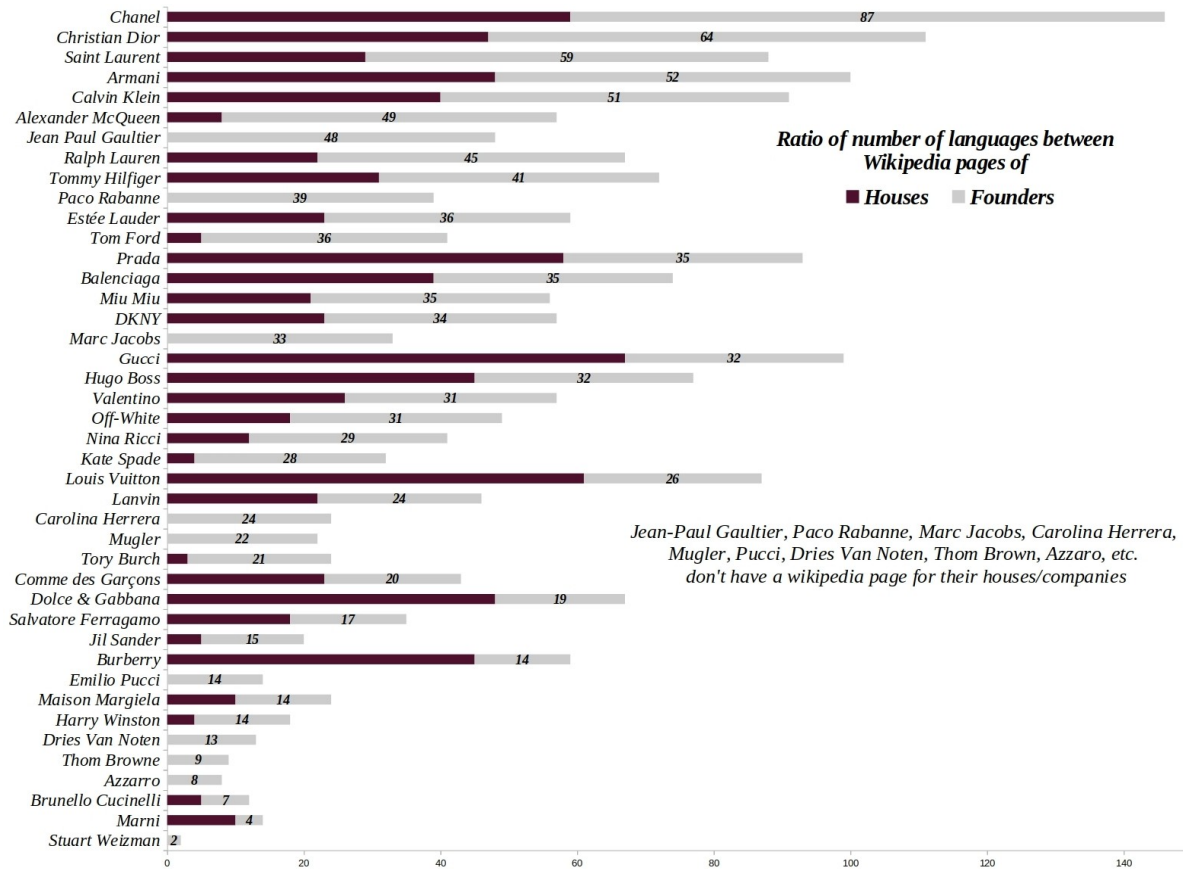

*Figure 4: Ratio between founders and houses on Wikipedia - Akim Mousterou, 2024*

Entirely focus on the creativity and without controlling interest, the "ArtisticDirector" have a non-linear vision by being inspired by the heritage of the House, while forecasting an aesthetic vision in their collections for the next seasons. The distinct creativity of an artistic director is interconnected and fundamentally conceived by repetition and difference (Deleuze, 1968).

# 6. Fine-tuned pretrained models

## 6.1 Bert, DistilBert, XML-R

Over the past decade, the NLP research community has been developing statistical language models leveraging deep learning architectures. The Transformer architecture and attention mechanism (Vaswani et al., 2017) opened the generation of pre-trained language models first encoder-only architecture, then encoder-decoder architecture, and then decoder-only architecture.

For the sequence labeling task, according to an input sequence of *n* tokens $x = (x_1, x_2, ..., x_n)$ the NER goal is to assign each token a label $y_i \in Y$ to indicate either the token is a part of a named entity (such as Date, Brand, Location, Private Company, Artistic Director, etc.) as $Y$ or do not belong to any entities (denoted as O class).

BERT (Devlin et al., 2018), which stands for Bidirectional Encoder Representations from Transformers stands defacto as a robust model to fine-tuned for downstream tasks. Lighter and faster at inference time, DistillBERT (Sanh et al., 2020) is a distilled version of BERT, leveraging a compression technique called, Knowledge distillation. XLM-R (Conneau et al., 2020) developed through the prism of multilingualism, constitute a powerful model for fine-tuned an English dataset with many French and Italian words.

Using open-source tools developed by ***Hugging Face***, we fine-tuned various NER models, first a BERT model, a DistilBERT model, and a XLM-RoBERTa model. During our annotation work, we observed that accents, diacritics, spaces, special characters, and letter case are key indicators of entities in the fashion and luxury industry. Those observations give an edge to XLM-R based on a SentencePiece tokenizer, built as a language-agnostic subword tokenizer at the Unicode characters level, with two segmentation algorithms, byte-pair-encoding (BPE) (Sennrich et al., 2016) and unigram language model (Kudo, 2018). Rather than BERT and DistilBERT models based on WordPiece tokenizer (Schuster and Nakajima, 2012), built as a purely data-driven subword-based tokenization with a custom greedy algorithm.

The purpose of the paper is to create a NLP framework for the fashion and luxury industry, we did not optimize hyper-parameters. We fine-tuned for 10 epochs according to basic hyper-parameters, with a learning rate of $5 \times 10^{-5}$, a training batch size of 32, and a validation batch size of 32, with an Adam optimizer (betas=(0.9,0.999) and epsilon=1e-08, with a linear scheduler.

## 6.2 Optimization strategies

To improve the efficiency, we implemented 3 applied NLP strategies. First, for the non-English rare words but popular in the luxury industry, we increased the size of the vocabulary by directly adding 300 new tokens in the custom tokenizer, and resized the token embedding matrix of the model so that its embedding matrix matches the tokenizer. Second, we alternate between cased and uncased for the BERT models. The cased model keeps the same text in the original papers as input, including both the capitalized and lowercase words, while the uncased models only use the words in lowercase. And thirdly, we leverage the method "*Class weighting*" to reduce the curse of the imbalanced classes in our real-world dataset. We assigned a manual rescaling weight in the cross-entropy loss that provides higher weights to the samples of the minority classes, and lower weights to the majority classes during the training process.

## 6.3 NER Luxury models fine-tuning benchmarks

| Model | Parameters | Size | Training Loss | Epoch | Step | Validation Loss | Precision | Recall | F1 | Accuracy |
|---|---|---|---|---|---|---|---|---|---|---|
| BERT (cased) | 108M | 432MB | 0.1024 | 8.0 | 9232 | 0.3519 | 0.7959 | 0.8377 | 0.8163 | 0.9497 |
| BERT (uncased) | 109M | 437MB | 0.1684 | 8.0 | 9232 | 0.4487 | 0.7441 | 0.7938 | 0.7682 | 0.9347 |
| DistilBert | 65.5M | 262MB | 0.1381 | 8.0 | 9240 | 0.3961 | 0.7613 | 0.8081 | 0.7840 | 0.9406 |
| XLM-R (Base) | 278M | 1,11G | 0.197 | 8.0 | 9240 | 0.3939 | 0.7638 | 0.8039 | 0.7833 | 0.9403 |
| XLM-R (Large) | 559M | 2,25 G | 0.0744 | 8.0 | 9240 | 0.3719 | **0.8176** | **0.8433** | **0.8302** | **0.9528** |

Conceptually simple and empirically powerful, BERT gave us top-notch results. Using BERT architecture coupled with a compression technique known as Knowledge distillation, DistilBERT gives us a robust model at lightweight size, easy scalability, and high-speed performance. While XLM-R gave us the best results but 4 times larger than the previous model, although able to potentially conduct cross-lingual transfer tasks. Due to the large number of labels, we ignored the accuracy metrics, to focus on the F1 score.

# 7. Benchmark NER models against Llama 3.1 Herd

## 7.1 Letter to Shareholder's, earnings call transcripts, and annual reports

Despite the strong general capabilities of LLMs, many papers noted that domain-specific NER tasks often perform better with supervised learning models, than with current LLMs (Qin et al., 2023). More recent works have shown that general-purpose LLMs can achieve comparable performance to fine-tuned models in standard benchmarks (Xie et al., 2023). However, their efficacy in our domain-specific scenarios has never been studied.

To ensure fairness and measure the performance, we compare our NER models and various LLMs, by using multiple letters from chairpersons to shareholders, excerpts of annual reports, by-laws, and transcripts of general assembly meetings, that are publicly available, and not present in our dataset to avoid data contamination. All the 50 paragraphs of the benchmark are available in the appendix. Below, the inference of Ner-Luxury model on the paragraph number 17 about M&A activities of the *Kering Group* as an example.

*Figure 5: #17 Kering - 2023 Universal Registration Document - Page 40*

## 7.2 Llama 3.1 Models

We leverage 3 fully open-source models Llama 3.1 in 8B-Instruct, Llama 3.1 70B-Instruct, and Llama 3.1 405B-Instruct (Dubey et al., 2024) from Meta Inc, powered on the enterprise-scale AI platform *SambaNova Systems Inc.* and compare them with Zero-shot and Few-shot methods.

|  | 8B | 70B | 405B |
|---|---|---|---|
| Layers | 32 | 80 | 126 |
| Model Dimension | 4,096 | 8192 | 16,384 |
| FFN Dimension | 14,336 | 28,672 | 53,248 |
| Attention Heads | 32 | 64 | 128 |
| Key/Value Heads | 8 | 8 | 8 |
| Peak Learning Rate | $3 \times 10^{-4}$ | $1.5 \times 10^{-4}$ | $8 \times 10^{-5}$ |
| Activation Function | SwiGLU | | |
| Vocabulary Size | 128,000 | | |
| Positional Embeddings | RoPE ($\theta = 500,000$) | | |

*Figure 6: Key hyperparameters of Llama 3.1 - Meta. July 23, 2024 [arXiv 2407.21783v2]*

## 7.3 Prompt template, zero-shot, and few-shot method

### - Zero-shot learning

For Zero-shot learning, the prompt template is composed with a list of entities with specific labels and a task description with a strict scope to improve reliability.

> Given entity label set: ['Location', 'Event', 'Monetary Value', 'Date', 'House', 'Brand', 'Fast Fashion', 'Private Company', 'Listed Group', 'Holding Trust', 'Investment Firm', 'Media Publisher', 'Hospitality', 'Museum Gallery', 'Retailer', 'Education', 'Organization', 'Artistic Director', 'Executive', 'Founder', 'Chairperson', 'Analyst Banker', 'KOL', 'Athlete Team', 'Model', 'Creative Insider', 'Editor Journalist', 'Garment Collection', 'Cosmetic', 'Fragrance', 'Bag Travel Goods', 'Jewelry', 'Timepiece', 'Footwear', 'Wine Spirit', 'Sustainability', 'Cultural Artifact'].
>
> Please recognize all the named entities in the given text. Based only on the given entity label set, provide answer in the following JSON format: [{'Entity Name': 'Entity Label'}]. If there is no entity in the text, return the following empty list: []."

### - Few-shot method

For the few-shot learning, we appended to the previous prompt, the following sentence "*Please found below four example:*" and 4 examples annotated with gold labels.

French house [Hermès HOUSE] and British department store [Selfridges RETAILER] are leaving the [Fashion Pact SUSTAINABILITY] after the appointment of CEO [Helena Helmersson EXECUTIVE] from Swedish fast-fashion company [H&M LISTEDGROUP] as the new co-chair

On [3 April 2023, DATE] [L'Oréal LISTEDGROUP] acquired for [$2.5 Billion MONETARYVALUE] the cosmetic label [Aēsop PRIVATECOMPANY] from [Australia. LOCATION] And on [26 June 2023, DATE] the French luxury group [Kering LISTEDGROUP] acquired 100% of the perfume house, [Creed PRIVATECOMPANY] from a fund of [BlackRock Inc. INVESTMENTFIRM]

During [Milan Fashion Week, EVENT] [Raf Simons ARTISTICDIRECTOR] and [Miuccia Prada ARTISTICDIRECTOR] showcased their latest [Prada HOUSE] collection at the [Fondazione Prada MUSEUMGALLERY] in [Milan. LOCATION]

According to [Bloomberg, MEDIAPUBLISHER] the market cap of [LVMH LISTEDGROUP] surpassed [$500 billion MONETARYVALUE] becoming the first European company to reach that milestone. As of [July 2023, DATE] [Hermès LISTEDGROUP] has a market cap of [$213.80 Billion, MONETARYVALUE] bigger than [Nike LISTEDGROUP] at [$161.80 Billion MONETARYVALUE]

*7.4 Luxury benchmark metrics NER models and Llama 3.1 models*

| Model | Precision | Recall | F1 |
|---|---|---|---|
| Ner-Luxury BERT (cased) | 0.9154 | 0.9114 | 0.9134 |
| Ner-Luxury XLM-R (Large) | **0.9530** | **0.9488** | **0.9509** |
| Llama 3.1 8B Zero-shot | 0.5039 | 0.3746 | 0.4298 |
| Llama 3.1 8B Few-shot | 0.7074 | 0.5991 | 0.6488 |
| Llama 3.1 70B Zero-shot | 0.6890 | 0.5910 | 0.6362 |
| Llama 3.1 70B Few-shot | 0.7810 | 0.7184 | 0.7484 |
| Llama 3.1 405B Zero-shot | 0.6882 | 0.6706 | 0.6793 |
| Llama 3.1 405B Few-shot | 0.8954 | 0.8456 | 0.8698 |

## 8. Models observations

### 8.1 NER-Luxury models

The main mistakes of our NER models are based on entities not seen during fine-tuning (Out-of-vocabulary – OOV), and boundaries of entity composed of multiple tokens. NER-Luxury struggles with specific semantic variations of legal and financial information. For example, the *S.E.C* filing nomenclature to introduce a listed-group with parentheses and quotation marks such as *Capri Holdings Limited ("Capri")*.

### 8.2 Zero-Shot with Llama 3.1

While, our proprietary dataset is updated on a monthly basis (last update in September 2024), we observed that the many mistakes made by Llama 3.1 were only due to the knowledge cutoff of December 2023. For example, the Spanish luxury group *Puig* was identified as "Private Company" but should be a "Listed Group" because *Puig* debuted their IPO on May 3rd, 2024.

The smallest model, Llama 3.1 8B-Instruct model encountered numerous difficulties to integrate a specific annotation scheme for the classic NER task. The Llama 3.1 8B-Instruct model predicted only "Brand" as a general group instead of various labels available "House", "Brand", "Fast Fashion", "Private Company", etc. It also ignores the suffix "S.p.A." in Italian or "S.A" in French, indicators of corporate structures. We also observed the difficulty to recognize entity of complex date, months or quarter alone.

Boosted by knowledge classification and scaling law experiments Llama 3.1 70B-Instruct model and Llama 3.1 405B-Instruct model, we can see the impressive knowledge extension into fashion and luxury domain with specific entity recognition of Asian retailers, fragrance names, sailors of the *America's Cup*, and upcoming celebrities.

Llama 3.1 70B-Instruct model and Llama 3.1 405B-Instruct model can perfectly separate "Brand" from "Company", yet our hierarchical classification task between luxury houses, brands, fast-fashion brands isn't properly achieved.

Due to the lack of transparency in the training data, we can only hypothesize that various European luxury groups carefully avoid the English words "Brand" or "Luxury brand" and prioritize the usage of "House" or even the French word "Maison". The subtle semantic trick enable them a *time-space distanciation* (Giddens, 1981), to be sublimated as a "House" and avoid the common classification of "Brand".

For example, on all Llama 3.1 models, Valentino is recognized as a "House" because *Valentino* is very often introduced as an "Italian couture House". On the contrary, *Tommy Hilfiger* is recognized as "Brand" because *Tommy Hilfiger* is mostly presented as "Luxury clothing brand" or as "Brand" by the parent-company, *PVH Corp.*

*8.3 Few-Shot with Llama 3.1*

When dealing with infrequent entity types in specific domains, LLMs may struggle to accurately extract entities using the universal prior knowledge acquired during pre-training. With only 4 examples in the few-shot method, we observed a drastic improvement of the NER task from all the Llama 3.1 models.

The three Llama 3.1 models are perfectly following the annotation rules of the few-shot method. Llama 3.1 70B-Instruct and 405B-Instruct models are generalizing more accurately on the classification task.

We observed that Llama 3.1 405B-Instruct is heavily penalized by finding extra entities by lemmatization words. For example, the model modified "European based operations" into "Europe" and attributed the label "Location", similarly "resumption of Chinese travel" is transformed into "China" for the label "Location". In future research, it would be wise to introduce a rule in the prompt template to regulate the information extraction via lemmatization to prevent unnecessary entities.

*8.4 Errors and improvements*

For better results on Llama 3.1, we revised a few labels from British English to American English. We also noted that the label "KOL (Key Opinion Leader)", our neutral label for musicians, singers, actresses, celebrities, artists, and historical figures that amplify signals might have been too obscure or with negative connotation.

For the few-shot method, our 4 examples might have been too narrow, covering 6 listed-groups, 3 luxury houses, and only 2 subdomains. By adding multiples distinct examples of various luxury groups, houses, domains, sub-sectors, we could further improve the overall metrics by leveraging the context length from 8K to 128K of the Llama 3.1 family.

# 9. Discussion & Limitations

Our literature review faces limitations, including potential biases due to the rare scientific analysis of the luxury industry, the highly unbalanced market with firms of unequal sizes, the divergence of activities or sub-sectors, which may affect the comprehensiveness of our analysis.

The encyclopedic nature of Wikipedia introduces a bias towards historical facts, potentially overlooking more recent events, giving an edge to century-old fashion houses compared to contemporary fashion houses. During the annotation, we observed that multiple small to mid-size luxury houses have historical inaccuracies, translation mistakes, and syntax inconsistencies, that unfortunately increase ambiguity, and induce suboptimal annotation workflows.

In Wikipedia, the sustainability aspect are often presented in a negative context in the "controversies" chapter and introduces negative biases. For the fast-fashion *Shein*, more than 50% of the English Wikipedia page are controversies (e.g, Taxes evasion, environmental impact, unsafe chemical exposure, Xinjiang Cotton). We could diminish this negative bias on sustainability by incorporating sentences from ESG reports of luxury groups that often introduce sustainability best practices (e.g, Greenhouse gas emissions reduction, carbon-neutral, regenerative agriculture, etc.).

We acknowledge that English annual reports are translated from Italian and French to facilitate the efficient communication with international shareholders. While the translation enrich the breadth and depth of the luxury houses part of listed companies, it also introducing various systematic, random, and human errors.

In a multilingual prism, entities in French and Italian are pretty close to English entities. For non-Indo-European languages, NER will bring complexity with multiple reading for example in Japanese, the French luxury house "*Louis Vuitton*" can be written using the romanization, "Louis Vuitton", in uppercase "LOUIS VUITTON", in hiragana for the formal way "ルイ・ヴィトン", and "ルイヴィトン" without the interpunct (U+30FB ・) in Unicode. Similarly, in Chinese, we can write with the romanization, "Louis Vuitton" and the transliterated version "路易威登" that closely mimic the pronunciation of the luxury house and include Chinese characters of positive connotations.

## 10. Conclusion and future works

Our research demonstrates the efficacy of using supervised learning in NLP for conducting named-entity recognition task for a specific domain, particularly within the rapidly evolving auto-regressive LLMs. For inference speed, our robust NER models only used CPU, on the contrary, Llama 3.1 requires multiple GPUs. However, as *Meta* encouraged synthetic data generation, we could leverage Llama 3.1 models in few-shot method for an automatic annotation workflow to increase the size of our dataset and reinforce the knowledge in various domain.

Regarding our NER dataset performance, there is an expected level of noise for NER datasets that we tried to reduce by enforcing multiple strategies. Our dataset of around 40K sentences might be smaller than many general purpose NER datasets, but is comparable in size to those in specialized fields. Nevertheless, the constant expansion of our dataset would likely improve its F1 score.

Our key contributions include the creation of a clear methodology, the construction of the first NER dataset focused on the fashion and luxury industry, the expansion towards beauty, sporting goods and fast-fashion and the implementation of various NER models.

By including financial reports of listed-groups in our annotation workflow, we could incorporate semantic and syntactic features of economic and financial terminology, therefore improving our overall metrics.
Additionally, we have conducted an inaugural empirical research using the methodology and dataset, demonstrating their practical benefits of disambiguation of entities, serving as valuable resources for researchers, policymakers, and industry experts.

Our analysis reveals the critical intersections of many topics. Moreover, this research represents a step toward classification and understanding of key mechanism of the fashion and luxury industry at scale with an open methodology, which offers a foundation for further research.

## Acknowledgments

We thank the leading enterprise-scale AI platform *SambaNova Systems Inc*. for the privileged access to their SambaNova Cloud API powered by their SN40L Chip. We would like to thank Clémentine Fourrier of *Hugging Face*, Pierre Mallevays of *Stanhope Capital*, Anthony Susevski of *RBC Capital Markets/Cohere*, Norbert Preining of *Arxiv*, Luca Solca of *Sanford C. Bernstein*, Toshinori Kujiraoka of *SambaNova*, Kaito Sugimoto of *NTT Communications*, Srikant Manas Kala of *Veritus AI*, Virginie Bray Esq., for their time, insightful comments, and constructive suggestions.

# Appendix

Full benchmark with 50 paragraphs and sources from luxury groups covering by order, *Ralph Lauren, Hermès, Burberry, Coty Inc., Salvatore Ferragamo, Kering, LVMH, Prada, Lanvin, Puig, Richemont, Swatch, Tapestry, Capri, Choi Tai Fook, Brunello Cuccinelli, Interparfums, L'Oréal, Chanel,* and *Moncler.*

## – Ralph Lauren Corporation –

**1.** *Ralph Lauren May 25, 2023 – Q4FY23 – Page 1*

"As I reflect on the past year, I am inspired by how our teams around the world brought the magic of our timeless vision to life. From our California [LOCATION] Dreaming [EVENT] show to sponsoring some of the most iconic moments in sports — it's their passion and optimism that inspire people to step into their dreams," said Ralph Lauren, [FOUNDER] Executive Chairman and Chief Creative Officer.

**2.** *Ralph Lauren May 25, 2023 – Q4FY23 – Page 1*

"We made strong progress in the first year of our Next Great Chapter: Accelerate [EVENT] plan, as our teams around the world executed exceptionally well through a highly dynamic global operating environment," said Patrice Louvet, [EXECUTIVE] President and Chief Executive Officer. "Our Fiscal [DATE] 2023 [DATE] performance puts us on track with our investor [EVENT] day commitments. We continue to be on offense as we balance growth and operating discipline, investing in our brand while delivering strong shareholder returns."

## – Hermès International S.C.A. –

**3.** *Hermès Universal Registration Document 2023 – Page 5*

2023 [DATE] was a busy year for events and opportunities to get together. In keeping with our tradition, we welcomed more than 52,000 visitors to the Hermès in the Making [EVENT] exhibition in Lille [LOCATION] (France), Chicago [LOCATION] (United States) [LOCATION] and Bangkok [LOCATION] (Thailand), and more than 35,000 visitors to the immersive La Fabrique de la légèreté [EVENT] exhibition in Taiwan, [LOCATION] Los Angeles [LOCATION] (United States), [LOCATION] Hong Kong [LOCATION] and Shanghai [LOCATION] (China).

**4.** *Hermès Universal Registration Document 2023 – Page 6*

Axel Dumas, [CHAIRPERSON] Executive Chairman of Hermès, [LISTEDGROUP] said: "In 2023, [DATE] Hermès [HOUSE] has once again cultivated its singularity and achieved an outstanding performance in all métiers and across all regions against a high base. These solid results reflect the strong desirability of our collections and the commitment and talent of the house's women and men. I thank them all warmly."

**5.** *Hermès Universal Registration Document 2023 – Page 6*

In 2023, [DATE] Hermès [LISTEDGROUP] inaugurated the leather goods workshops in Louviers [LOCATION] and la Sormonne, [LOCATION] the first two industrial buildings in France [LOCATION] to carry the E4C2 [SUSTAINABILITY] label that assesses environmental performance based on energy consumption and carbon emissions. [SUSTAINABILITY]

### 6. Hermès Universal Registration Document 2023 – Page 6

Asia excluding Japan (+19%) pursued its strong growth, with significant increases in sales in all the countries of the region. A second store opened in October in Chengdu, the capital city of the province of Sichuan, becoming the house's thirty-third address in Mainland China, following the opening of a store in Tianjin in July. In Korea, the store at the Shilla Hotel in Seoul reopened in December after renovation and extension work. Japan (+26%) recorded a steady and sustained increase in sales. The Daimaru Sapporo store on Hokkaido island and the Takashimaya store in the centre of Kyoto were inaugurated in October and November, after renovation and expansion.

## – Burberry Group Plc, –

### 7. Burberry Annual Report 2023/24 – Page 1

Burberry is a British luxury brand headquartered in London with a longstanding commitment to quality, innovation, creativity and responsible business. Our brand is built on the principles of Thomas Burberry, who founded the Company in 1856. With his invention of gabardine in 1879, Thomas revolutionised outerwear and opened opportunities for adventurers to explore new spaces. We continue that legacy today as we focus on bringing our vision of Modern British Luxury to life.

### 8. Burberry Annual Report 2023/24 – Page 4

There have been a number of Board changes during FY 2023/24. On behalf of the Board, I would like to thank Matthew Key, who retired from the Board on 12 July 2023, for his service to Burberry, including as Audit Committee Chair. We welcomed Kate Ferry, who joined the Board as Chief Financial Officer on 17 July 2023. Kate joined us from McLaren Group where, as Chief Financial Officer, she oversaw financial strategy and investor relations. It was also my pleasure to welcome Alessandra Cozzani who joined the Board as an independent Non-Executive Director on 1 September 2023. Alessandra previously served as Chief Financial Officer of Prada Group SpA and her financial and luxury fashion expertise make her a valued addition to our Board. Further information on Board recruitment and the induction processes for Kate and Alessandra is provided in the Nomination Committee Report on pages 113 to 117.

### 9. Burberry Annual Report 2023/24 – Page 19

During the year, we have focused on what makes Burberry unique, reinforcing our connection with Britishness and our heritage of the outdoors. We held our Summer 2024 show in a tent in Highbury Fields during London Fashion Week in September 2023. The event helped amplify our brand visibility and received positive press responses as well as good customer engagement. Our Spring 2024 campaign focused on discovering London's urban landscape, underscoring our deep connection with the city.

*– Coty Inc. –*

**10.** *Coty Annual Report 2023/24 – Page 36*

Net revenues grew across both our segments. The growth in our Consumer Beauty segment was due to positive performance across the body care, skincare, and color cosmetics categories. Growth in our Prestige segment was primarily due to the positive performance in the prestige fragrance category due to the continued success of fragrance brands such as Burberry, Calvin Klein, Hugo Boss, Gucci, and Marc Jacobs. Although, the prestige makeup category was negatively impacted by COVID-19 related to the lockdowns in China in the earlier portion of the fiscal period, this category began to show recovery in the last quarter of the fiscal period. The overall increase in net revenues reflects the successful implementation of global price increases across all product categories, our product premiumization strategy, and positive overall market trends.

**11.** *Coty Annual Report 2023/24 – Page 37*

(i) the continued success and growth of prestige fragrances, specifically Burberry Hero, Burberry Her, Calvin Klein Hugo Boss Boss Bottled, Gucci Flora, and Marc Jacobs Daisy; (ii) the positive pricing impact as a result of global price increases and in line with the overall premiumization strategy; (iii) growth in travel retail net revenues in all major regions due to increased leisure travel compared to the prior year; and (iv) growth in the U.S due to positive market trends and innovation in the prestige fragrance brands. These increases were partially offset by: (i) lower net revenues in the Prestige makeup category impacted by a decline in Gucci makeup travel retail sales in the Asia Pacific region as a result of slow recovery from the lockdowns in China; and (ii) lower net revenues for philosophy due to less innovation and repositioning of the brand.

*– Salvatore Ferragamo S.p.A. –*

**12.** *Ferragamo Annual report 2023 – 31 December 2023 – Page 5*

In 2023, our Group continued the development process begun last year, inspired by its history, especially its founding values, its Italian and craftsmanship roots, and its international vocation that finds its energy precisely in its close ties with Florence.

**13.** *Ferragamo Annual report 2023 – 31 December 2023 – Page 5*

In strengthening Ferragamo's identity in the vision of the future, we also found important confirmation in the appreciation of the new collections and the new creative direction, coming also from the important awards given to Maximilian Davis, our talented young Creative Director. In fact, Maximilian won the prestigious Best Designer of the Year award for womenswear from The Fashion Awards and was named Men of the Year by GQ Italia.

***14.** Ferragamo Annual report 2023 – 31 December 2023 – Page 5*

2023 is the year we improved our ESG ratings and updated our Sustainability Plan, setting important new environmental, social and governance goals. Creating, re-imagining, being style icons and innovating, while remaining true to our core values: Innovation, Creativity, Craftsmanship, Sustainability, Authenticity, Independence, and the Italian spirit are the basis of our development process, which will continue and strengthen in 2024 and the years to come. For all of this, I would like to thank the entire team led by Marco Gobbetti who, with love, dedication, tenacity, and enthusiasm, has accepted and embraced the challenge of continuously developing our brand, in Italy and around the world, so that Ferragamo will continue to be the "Dream brand". And of course, and most importantly, I would like to thank our many customers who appreciate our values and products and who are and will always be at the center of everything we do. We are grateful to them for placing their trust in a brand that will always want to put them at the center, because of our great and constant commitment to always want to express the best, to represent the best that our beloved Italy has to offer! Leonardo Ferragamo Salvatore Ferragamo S.p.A. Chair

*– Kering S.A –*

***15.** Kering Universal Registration Document 2023 – Page 16*

Kering acquired a 30% shareholding in Italian couture House Valentino for €1.7 billion as part of a strategic partnership with investment company Mayhoola. The agreement comprises an option for Kering to acquire 100% of Valentino's share capital no later than 2028. Appointment of Raffaella Cornaggia as Chief Executive Officer of Kering Beauté and as a Group Executive Committee member. Supported by a team of seasoned professionals, her role is to develop the expertise of Bottega Veneta, Balenciaga, Alexander McQueen, Pomellato and Qeelin in the Beauty category, while also unlocking the full potential of luxury fragrance House Creed, which was acquired in 2023. This acquisition has given Kering Beauté a platform to support the future development of other fragrances.

***16.** Kering Universal Registration Document 2023 – Page 16*

Kering is the result of one family's entrepreneurial journey. It is 42.2%-owned by the Artémis holding company, which is controlled by the Group's founding Pinault family. Alongside this solid core shareholder, the Group's ownership structure has become increasingly international over a period of more than 10 years, reflecting the Group's worldwide growth and transformation. Institutional investors own 52.7% of the Group's capital.

**17.** *Kering Universal Registration Document 2023 – Page 40*

This category has estimated revenue of €56-72 billion. On February 3, 2023, Kering announced the creation of Kering Beauté in order to develop its Beauty expertise for Bottega Veneta, Balenciaga, Alexander McQueen, Pomellato and Qeelin. Beauty is a natural extension of Kering's Luxury universe. On June 26, 2023, Kering also announced the acquisition of Creed, an established high-end luxury fragrance House. The high-end luxury fragrance segment was worth an estimated €5 billion in 2022 according to Bain and is growing rapidly, with growth expected to average 15% per year between 2022 and 2026. Kering also operates in the Perfumes & Cosmetics category through licensing agreements with certain leading industry players, such as L'Oréal, Coty and Interparfums, for its Saint Laurent, Gucci and Boucheron brands.

### – LVMH Moët Hennessy Louis Vuitton SE –

**18.** *LVMH Universal Registration Document 2023 – 31 December 2023 – Page 40*

Our performance in 2023 illustrates the exceptional appeal of our Maisons and their ability to spark desire, despite a year affected by economic and geopolitical challenges. The Group once again recorded significant growth in revenue and profits. Our growth strategy, based on the complementary nature of our businesses, as well as their geographic diversity, encourages innovation, high-quality design and retail excellence, and adds a cultural and historical dimension thanks to the heritage of our Maisons. This was reflected in Louis Vuitton and Christian Dior's spectacular fashion shows, Tiffany's reopening of "The Landmark" in New York and the ever growing popularity of Sephora's store concept worldwide. 2023 also saw us make progress in several key areas that are essential components of our long-term vision: protecting the environment, developing our talent, and preserving and passing on our expertise. While remaining vigilant in the current context, we enter 2024 with confidence, backed by our highly desirable brands and our agile teams. It promises to be an inspiring, exceptional year for us all, featuring our partnership with the Paris 2024 Olympic and Paralympic Games, whose core values of passion, inclusion and surpassing oneself are shared by our Group. For LVMH, it provides a new opportunity to reinforce our global leadership position in luxury goods and promote France's reputation for excellence around the world.

Bernard Arnault Chairman and Chief Executive Officer January 2024

– *Prada S.p.A* –

**19.** *Prada annual report – 31 December 2024 – Page 8*

Let me start on a personal note, by saying that this first year at the Prada Group [LISTEDGROUP] marks the beginning of an extremely fulfilling chapter for me and that I am extremely proud of being a part of this incredible journey. 2023 [DATE] was a remarkable year, as we achieved our ambitions against the backdrop of increasing macroeconomic and geopolitical uncertainty, especially in the second half of the year. [DATE] The Group generated revenue of Euro 4.7bn, [MONETARYVALUE] with growth up +17%. The strong performance of the fourth quarter [DATE] (+17%), marks the 12th consecutive quarter [DATE] of growth. At a brand level, Prada's [LISTEDGROUP] retail sales grew at a solid +12%, while Miu Miu [HOUSE] thrived with +58%. Alongside these excellent topline results, we also improved our profitability, reaching a 22.5% EBIT margin, which also reflects significant investments behind our brands. Andrea Guerra [CHAIRPERSON] Chief Executive Officer and Executive Director

**20.** *Prada annual report – 31 December 2024 – Page 13*

The Prada Group [LISTEDGROUP] is a global leader in the luxury industry and a pioneer in its unconventional dialogue with contemporary society across diverse cultural spheres. Prada Group [LISTEDGROUP] also operates in the eyewear and beauty sectors through licensing agreements with industry leaders. Prada S.p.A. [LISTEDGROUP] is listed on the Hong Kong Stock Exchange [PRIVATECOMPANY] as 1913. [LISTEDGROUP]

**21.** *Prada annual report – 31 December 2024 – Page 24*

Since 1913 [DATE], Prada [HOUSE] has been synonymous with cutting-edge style. Its intellectual universe combines concept, structure and image through codes that go beyond trends. Its fashion transcends products, translating conceptuality into a universe that has become a benchmark to those who dare to challenge conventions. With its collections, today [DATE] Prada [HOUSE] embodies and spreads the vision and intellectual curiosity of co-creative directors Miuccia Prada [ARTISTICDIRECTOR] and Raf Simons [ARTISTICDIRECTOR].

– *Lanvin Group* –

**22.** *Lanvin Investor Relations landing page – Corporate Overview*

Founded in 2017, [DATE] Lanvin Group [LISTEDGROUP] is the leading global luxury fashion group headquartered in China, [LOCATION] managing iconic brands worldwide including the oldest operating French couture house Lanvin [HOUSE] founded in 1889, [DATE] Italian luxury shoemaker Sergio Rossi, [HOUSE] Austrian skinwear specialist Wolford, [HOUSE] iconic American womenswear brand St. John Knits, [HOUSE] and high-end Italian menswear maker Caruso. [HOUSE] With over 390 years of combined history, these five brands have far-reaching global presence, operating in more than 80 countries with approximately 1,200 points of sales, 3,600 employees and over 300 retail stores across the world.

**23.** *Puig annual report 2023 – 31 December 2023 – Page 17*

Puig closed 2023 with net revenue of €4,304 million (+19% vs 2022), EBITDA of €849 million, and net profit of €465 million. These results will allow a dividend to be paid in the first half of 2024 of €186 million. If we look back, we can say proudly that these are the best results in Puig history, but I would like to put this data into perspective to really understand the scope of these numbers. In 2021, we said we were going to double our 2020 sales in three years (which meant reaching €3 billion by the end of 2023) and we would triple them in five years (€4.5 billion by the end of 2025). The reality is that we doubled sales in two years instead of three, and we finished 2023 not far off the results we had expected to achieve in 2025.

**24.** *Puig annual report 2023 – 31 December 2023 – Page 18*

In addition, designer Dries Van Noten, whose eponymous brand joined our portfolio in 2018, was honored as Designer of the Year for his unique and colorful approach not only to his garments, but also to his accessories and beauty line. We have closed the third and final year of the 2021-23 strategic plan working with a well-established organizational structure, which was strengthened with the addition of Marc Toulemonde as President of Derma, and with the appointment of Marine de Boucaud as Puig Deputy Corporate Human Resources Officer, and who succeeds Eulalia Alfonso as Chief Human Resources Officer on January 1, 2024.

**25.** *Puig annual report 2023 – 31 December 2023 – Page 18*

Our fragrance and fashion segment posted a 17% increase in net revenue and gained market share thanks to the good performance of the iconic 1 Million from Rabanne and Good Girl from Carolina Herrera, and the successful launches of Jean Paul Gaultier's Le Male Elixir and Gaultier Divine. Rabanne has been the first of our brands to reach the €1 billion mark in revenue, and Jean Paul Gaultier is the brand that has experienced most growth this year.

– *Compagnie Financière Richemont S.A.* –

**26.** *Richemont annual report 2023 – 31 March 2023 – Page 2*

All business areas delivered double-digit sales growth compared to the prior year. Our Jewellery Maisons, Buccellati, Cartier and Van Cleef & Arpels, increased their combined sales to €13.4 billion and operating profit to €4.7 billion, generating an improved 34.9% operating margin compared to the prior year. While Buccellati continued to develop solidly, generating the highest growth rates across the Group albeit from a smaller base, Cartier and Van Cleef & Arpels reaffirmed their market leadership with a high level of sales growth and profitability. The Jewellery Maisons enjoy the highest level of direct-to-client engagement within the Group (83%).

**27.** *Richemont annual report 2023 – 31 March 2023 – Page 3*

At the 2023 AGM, shareholders will be asked to elect two new directors to the Board: Fiona Druckenmiller and Bram Schot. Ms Druckenmiller's jewellery expertise, understanding of the American clientele and social and environmental causes will be of great value to the Board, while Mr Schot brings more than three decades of experience in the premium automotive industry and a deep understanding of risk management, supply chain and sustainability issues.

**28.** *Richemont annual report 2023 – 31 March 2023 – Page 4*

Economic volatility and political uncertainty look set to remain features of the trading environment. The Group will therefore seek to maintain the necessary agility to manage fluctuating levels of demand. I am confident that our Maisons are well positioned to meet strong demand, notably driven by a significant resumption of Chinese travel. Richemont is fortunate to own such a unique portfolio of Maisons with excellent long-term prospects. Johann Rupert Chairman Compagnie Financière Richemont SA

**29.** *Richemont annual report 2023 – 31 March 2023 – Page 3*

Acknowledging the need to embed sustainability even more firmly in our governance and reinforce the integral nature and importance of this discipline, we nominated Mr Schot to the Board and appointed Dr Ruchat to the SEC. This year, further comprehensive change was also initiated across Group functions, regions and Maisons to fully integrate ESG principles into all Richemont strategic and operational decision-making processes.

– *The Swatch Group Ltd* –

**30.** *Swatch annual report 2023 – 31 December 2023 – Page 17*

In its second orbit, the MoonSwatch continued to captivate the world, maintaining stellar levels of interest. The year's Mission to MoonshineTM Gold events were a celestial spectacle, drawing large crowds at each point of sale. Every month since March, the Mission to the Moon model featured a unique seconds hand that was inspired by the magic of each full moon and made of Omega's MoonshineTM Gold.

**31.** *Swatch annual report 2023 – 31 December 2023 – Page 24*

The sparkling Aegean Sea [LOCATION] was the destination of choice for Omega's [HOUSE] Seamaster [TIMEPIECE] celebration in summer of 2023. [DATE] Hosted in Mykonos, [LOCATION] it was a chance to unveil the brand's 11 new watches in Summer [GARMCOLLECTION] Blue. [EVENT] The event welcomed special guests including George Clooney [KOL] and Naomie Harris, [KOL] as well as the explorer Victor Vescovo, [KOL] and sailors Blair Tuke [KOL] and Peter Burling. [KOL]

### – *Tapestry Inc.,* –

**32.** *Tapestry form 10K – July 1, 2023 – Page 2*

Founded in 1941, [DATE] Coach, Inc., [LISTEDGROUP] the predecessor to Tapestry, Inc. [LISTEDGROUP] (the "Company"), was incorporated in the state of Maryland [LOCATION] in 2000. [DATE] During fiscal 2015, [DATE] the Company acquired Stuart Weitzman Holdings LLC, [PRIVATECOMPANY] a luxury women's footwear company. During fiscal 2018, [DATE] the Company acquired Kate Spade & Company, [PRIVATECOMPANY] a lifestyle accessories and ready-to-wear [GARMCOLLECTION] company. Later in fiscal 2018, [DATE] the Company changed its name to Tapestry, Inc. [LISTEDGROUP]

**33.** *Tapestry form 10K – July 1, 2023 – Page 20*

Tapestry, Inc. [LISTEDGROUP] is a New York-based [LOCATION] house of iconic accessories and lifestyle brands. Our global house of brands unites the magic of Coach, [HOUSE] kate spade new york [HOUSE] and Stuart Weitzman. [HOUSE] Each of our brands are unique and independent, while sharing a commitment to innovation and authenticity defined by distinctive products and differentiated customer experiences across channels and geographies.

### – *Capri Holding Limited* –

**34.** *Capri form 10K – July 1, 2023 – Page 2*

The Covid-19 pandemic [EVENT] has resulted in varying degrees of business disruption for the Company since it began in fiscal 2020 [DATE] and has impacted all regions around the world, resulting in restrictions and shutdowns implemented by national, state, and local authorities. Such disruptions continued during the first half of [DATE] fiscal 2023, [DATE] and the Company's results in Greater China [LOCATION] (mainland China, [LOCATION] Hong Kong SAR, [LOCATION] Macao SAR, [LOCATION] and Taiwan) [LOCATION] were adversely impacted as a result of the Covid-19 pandemic. [EVENT] Capri Holdings Limited [LISTEDGROUP] ("Capri") is a global fashion luxury group consisting of iconic, founder-led brands Versace, [HOUSE] Jimmy Choo [HOUSE] and Michael Kors. [HOUSE] Our commitment to glamorous style and craftsmanship is at the heart of each of our luxury brands.

*– Chow Tai Fook Jewellery Group Limited –*

**35.** *Chow Tai Fook annual report – March 31, 2024 – Page 2*

"As Chow Tai Fook Jewellery celebrates 95 years of excellence, we honour our legacy while looking steadfastly towards the future. We are in the process of writing an exciting new chapter in our brand's story, guided by our vision: To be the leading global jewellery brand that is a trusted lifetime partner for every generation." Dr. Cheng Ka-Shun, Henry Chairman

**36.** *Chow Tai Fook annual report – March 31, 2024 – Page 13*

During FY2024, the Group successfully grew its revenue by 14.8% to HK$108,713 million. Thanks to our disciplined approach in cost management, COP increased by 28.9% to HK$12,163 million, which operating margin improved to 11.2%. This exceptional performance led to record-high levels for both revenue and COP. Profit attributable to shareholders grew by 20.7% to HK$6,499 million. Earnings per share was HK$0.65. The Board has proposed a final dividend of HK$0.30 per share, bringing the dividend per share for the year to HK$0.55. The full year payout ratio in FY2024 was approximately 84.6%.

*– Brunello Cucinelli S.p.A. –*

**37.** *Brunello Cucinelli annual report 2023 – December 31, 2023 – Page 4*

We were particularly honoured by two recognitions in 2023. The first, which concerns the taste and philosophy that distinguish us, is the 2023 Designer of the Year award, which we received from the men's magazine GQ China in the beautiful town of Puyuan. The second, which rewards the path we have taken together since we were listed on the Italian Stock Exchange, and the entry of our stock in the FTSE MIB, the main index of Piazza Affari, which is a result that makes us very proud, especially for the way in which we obtained it together in a little more than ten years. These recognitions are added to a series of prestigious awards that have recently testified to the international recognition of our distinctive style ad our way of doing business, starting with the illustrious 2021 Designer of the Year award received in London from British GQ. Furthermore, in October 2021 we had the great honour of being invited to the G20 in Rome, where we have the opportunity to present the model and ideal that guide our work to important politicians from around the world. Finally, we received the "Neiman Marcus Fashion Award" in 2022, a type of "Oscar" for the fashion world.

**38.** *Brunello Cucinelli annual report 2023 – December 31, 2023 – Page 4*

As a result of this fantastic path we have created together, I would like to say something to all the young people of humanity, who I suggest to safeguard and share a special feeling, that of feeling as being the "guardians of the creation". A deep feeling that is able to guide sensitivity towards people who need help, to indicate the way towards a correct balance between profit and giving back, to find a harmonious and fruitful balance between body, mind and soul. Solomeo, 14 March 2024 Brunello Cucinelli Chairman of the Board of Directors

## – Inter Parfums Inc, –

**39.** *Inter Parfums form 10K – December 31, 2023 – Page 2*

We operate in the fragrance business, and manufacture, market and distribute a wide array of prestige fragrances and fragrance related products. We manage our business in two based operations, our European based operations and our United States based operations. Certain prestige fragrance products are produced and marketed by our European based operations through our 72% owned subsidiary in Paris, Interparfums SA, which is also a publicly traded company as 28% of Interparfums SA shares trade on the Euronext.

**40.** *Inter Parfums form 10K – December 31, 2023 – Page 3*

In October 2021, we closed on a transaction agreement with Salvatore Ferragamo S.p.A., whereby an exclusive and worldwide license was granted for the production and distribution of Ferragamo brand perfumes. Our rights under this license are subject to certain minimum advertising expenditures and royalty payments as are customary in our industry. The license became effective in October 2021 and will last for 10 years with a 5-year optional term, subject to certain conditions.

**41.** *Inter Parfums form 10K – December 31, 2023 – Page 3*

In September 2021, we entered into a long-term global licensing agreement for the creation, development and distribution of fragrances and fragrance-related products under the Donna Karan and DKNY brands. Our rights under this license are subject to certain minimum advertising expenditures and royalty payments as are customary in our industry. With this agreement, we have gained several well-established and valuable fragrance franchises, most notably Donna Karan Cashmere Mist and DKNY Be Delicious, as well as a significant loyal consumer base around the world. In connection with the grant of license, we issued 65,342 shares of Inter Parfums, Inc. common stock valued at $5.0 million to the licensor. The exclusive license became effective on July 1, 2022, and we are planning to launch new fragrances under these brands in 2024.

## – L'Oreal S.A. –

**42.** *L'Oréal universal registration document 2023 – 31 December 2023 – Page 11*

(1) Consisting, in addition to Ms Françoise Bettencourt Meyers, of Mr Jean-Pierre Meyers, Mr Jean-Victor Meyers and Mr Nicolas Meyers, along with Téthys S.A.S. and Financière L'Arcouest S.A.S. (2) Concerns the current and former employees of L'Oréal. The percentage also includes the performance shares granted in accordance with Article L. 225-197-1 of the French Commercial Code. Of which 0.99% of the share capital as part of the L'Oréal Employee Savings Plan and employee investment funds as defined by Article L. 225-102 of the French Commercial Code.

**43.** *L'Oréal universal registration document 2023 – 31 December 2023 – Page 47*

On 30 August, L'Oréal announced the completion of the acquisition of Aēsop – a brand that is highly complementary to the Luxe portfolio and offers significant growth potential, notably in China and Travel Retail. • In August, the Musée du Louvre and Lancôme joined forces for an unprecedented collaboration. Inspired by nine masterpieces of sculpture, Lancôme imagined a new skincare and makeup collection, signed Lancôme x Louvre. The launch campaign was filmed at the Louvre with several ambassadors of the brand.

**44.** *L'Oréal universal registration document 2023 – 31 December 2023 – Page 58*

The Internal Rules of the Board stipulate the following duties of Mr Jean-Paul Agon, in his capacity as Chairman of the Board of Directors: "The Chairman of the Board of Directors organises and oversees the Board's work and reports there on to the Annual General Meeting.

### – Chanel Limited –

**45.** *Chanel annual report 2023 – 31 December 2023 – Page 1*

"The strong results we are announcing today reflect Chanel's relentless focus on exceptional creations that inspire. They underline sustained investment in building the desirability of our brand, creating the ultimate luxury experience for our clients and supporting our people to grow and develop. Our belief in the transformative role of creation, our desire to shape what's next and our long-term perspective guide our approach. In 2023, we increased headcount globally by 14% to more than 36,500 people, expanded our retail distribution network to over 600 boutiques worldwide, and invested significantly in R&D and in technology. We also launched our Open Innovation Function to partner with start-ups, thought leaders and academic institutions. Chanel continues to contribute positively to the environment and the communities in which we operate. Our House Sustainability Ambition outlines our focus on restoring nature and climate, investing in circularity, supporting those in our supply chain and promoting the autonomy of women. We are proud of our new Net-Zero 2040 targets validated by the SBTi, which will underpin the continued transformation of our business.

**46.** *Chanel annual report 2023 – 31 December 2023 – Page 2*

Fashion creations and collections continued to inspire our clients with exceptional growth across all categories, particularly in Ready-to-Wear, Leather Goods and Shoes. The Cruise collection was unveiled in Los Angeles, aligning with the reopening of the flagship boutique on Rodeo Drive, and shown later in the year in Shenzhen. The collection was inspired by the multiple facets of Los Angeles, where in 1931, Chanel began its long and storied history with the city, working with Sam Goldwyn on costume designs for Hollywood films. On the other side of the Atlantic in 2023, the Métiers d'art fashion show was held in Manchester, fortifying the century-old relationship between Chanel and the UK. The collection was designed in celebration of Gabrielle Chanel's art of tailoring.

*– Moncler S.p.A. –*

**47.** *Moncler FY 2023 Financial Results Conference Call – February 28, 2024 – Page 17*

MELANIA GRIPPO: [ANALYSTBANKER] Good evening, everyone. This is Melania Grippo [ANALYSTBANKER] from BNP Paribas. [INVESTMENTFIRM] I have 2 questions. First on retail, if you could please tell us which trends you are currently seeing in retail compared to what you reported in Q4, [DATE] if you have any significant changes in terms of geographies or clusters? And the second question is on the Chinese cluster specifically. Could you please tell us how did it do in Q4 [DATE] on a year-on-year basis and versus 2021? [DATE] And also, if you can give us an idea on how is doing now after the Chinese New Year. [EVENT] Thank you

**48.** *Moncler annual report 2023 – December 31, 2023 – Page 17*

For Moncler Grenoble, [HOUSE] 2023 [DATE] saw the launch of collections across the entire first half of the year (Spring-Summer 2023 [DATE] in February [DATE] and Pre-Fall 2023 [DATE] in June) [DATE] for the first time ever, presenting essential outdoor layers for all seasons on the mountains. In December 2023, [DATE] Moncler [HOUSE] opened – in the heart of St. Moritz [LOCATION] – its first-ever Moncler [RETAILER] Grenoble [LOCATION] flagship store entirely dedicated to the world of Moncler Grenoble [HOUSE] and at the same time launched the new global campaign "Beyond performance". The new campaign stars four world-renowned winter athletes – Xuetong Cai, [ATHLETETEAM] Perrine Laffont, [ATHLETETEAM] Richard Permin, [ATHLETETEAM] and Shaun White [ATHLETETEAM] – and captures the group as they commune with the mountain and each other, cutting from adrenalin-fuelled descents to more intimate, off-duty moments.

**49.** *Moncler annual report 2023 – December 31, 2023 – Page 35*

Moncler [LISTEDGROUP] 's market capitalisation was €15.3 billion [MONETARYVALUE] as at 31 December 2023 [DATE] , compared to €13.6 billion [MONETARYVALUE] as at 31 December 2022 [DATE] , and in the year recorded a Total Shareholder Return (TSR) of 14%. The number of shares was 274,627,673 as at 31 December 2023 [DATE] . Moncler [LISTEDGROUP] 's significant shareholders are shown in the chart below

23.7% Double R [INVESTMENTFIRM] S.r.l. [HOLDINGTRUST]
8.6% Morgan Stanley [INVESTMENTFIRM]
5.0% Capital Research and Management Company [INVESTMENTFIRM]
4.2% BlackRock Inc. [INVESTMENTFIRM]
1.6% of treasury shares
56.9% Other shareholders

**50.** *Moncler FY 2023 Financial Results Conference Call – February 28, 2024 – Page 33*

OPERATOR : The next question is from Louise Singlehurst [ANALYSTBANKER] from Goldman Sachs. [INVESTMENTFIRM] Please go ahead. LOUISE SINGLEHURST: [ANALYSTBANKER] Hi, good evening, everyone. Thanks for taking my questions. I wondered if I could just ask a little bit more about the US. [LOCATION] I'm quite interested in seeing there is a big improvement going from Q3 [DATE] into Q4 [DATE] on the DTC, excluding the wholesale. Do you think now we've turned the corner in the US [LOCATION] for more sustainable positive growth? I know, Roberto, [EXECUTIVE] you mentioned that all nationalities had started the year well. So obviously a positive implication so far. But I wondered if we're on track for more sustainable growth now in the US? [LOCATION]